\documentclass[letterpaper, 10 pt, conference]{ieeeconf}

\IEEEoverridecommandlockouts
\makeatletter
\let\NAT@parse\undefined
\makeatother
                                  
\usepackage{graphicx} 
\usepackage{pgf} 
\usepackage{import} 
\usepackage{algorithm} 
\usepackage{algpseudocode} 
\usepackage{amsmath} 
\usepackage{booktabs}
\usepackage{multirow}
\usepackage{float}
\usepackage[
  colorlinks=true,
  linkcolor=blue,
  citecolor=red,
  urlcolor=cyan,
  filecolor=magenta
]{hyperref}
\usepackage{xurl}
\usepackage{tabularx}
\usepackage{acronym}
\usepackage{siunitx}
\usepackage{newtxtext}

\hypersetup{
    pdftoolbar=true,        
    pdfmenubar=true,        
    pdffitwindow=false,     
    pdfstartview={FitH},    
    pdftitle={One-Shot Badminton Shuttle Detection for Mobile Robots},    
    pdfauthor={Florentin Dipner},     
    pdfsubject={IEEE RAP},   
    pdfcreator={Florentin Dipner},   
    pdfproducer={Florentin Dipner}, 
    colorlinks=true,
    linkcolor=blue,
    linkbordercolor=white,
    filecolor=magenta,
    citecolor=red,
    urlcolor=cyan
    }
\usepackage[nameinlink,noabbrev]{cleveref}
\newenvironment{leftitemize}
  {\begin{list}{$\bullet$}{%
      \setlength{\leftmargin}{0pt}%
      \setlength{\itemindent}{15pt}%
      \setlength{\labelwidth}{0pt}%
      \setlength{\labelsep}{5pt}%
      \setlength{\itemsep}{0.0pt}%
      \setlength{\topsep}{0.0pt}%
    }}
  {\end{list}}

\newenvironment{leftenumerate}
  {%
    \begin{list}{\arabic{enumi}.}{%
      \usecounter{enumi}%
      \setlength{\leftmargin}{0pt}%
      \setlength{\itemindent}{15pt}%
      \setlength{\labelwidth}{0pt}%
      \setlength{\labelsep}{5pt}%
      \setlength{\itemsep}{0.0pt}%
      \setlength{\topsep}{0.0pt}%
    }%
  }
  {%
    \end{list}%
  }

\acrodef{fps}[FPS]{Frames per Second}
\acrodef{roi}[ROI]{Region of Interest}
\acrodef{cnn}[CNN]{Convolutional Neural Network}
\acrodef{map}[mAP]{Mean Average Precision}
\acrodef{iou}[IoU]{Intersection over Union}
\acrodef{fov}[FOV]{Field of View}
\acrodef{gmm}[GMM]{Gaussian Mixture Model}
\acrodef{nms}[NMS]{Non-Maximum Suppression}
\acrodef{gpu}[GPU]{Graphics Processing Unit}
\acrodef{fp}[FP]{False Positive}
\acrodef{tp}[TP]{True Positive}
\acrodef{fn}[FN]{False Negative}
\acrodef{dpr}[DPR]{Distance Precision Rate}

\title{\LARGE \bf One-Shot Badminton Shuttle Detection for Mobile Robots}

\author{Florentin Dipner$^{\dagger}$, William Talbot$^{\dagger}$, Turcan Tuna$^{\dagger}$, Andrei Cramariuc$^{\dagger}$, Marco Hutter$^{\dagger}$
\thanks{$^{\dagger}$Authors are with Robotic Systems Lab (RSL), ETH Zurich, Zurich 8092, Switzerland. {\tt\small \{fdipner, wtalbot, tutuna, crandrei, mahutter\}@ethz.ch}}
\thanks{Corresponding Author: William Talbot, wtalbot@ethz.ch}
}

\begin{document}

\maketitle

\thispagestyle{empty} \pagestyle{empty}

\begin{abstract}
This paper presents a robust one-shot badminton shuttlecock detection framework for non-stationary robots. To address the lack of egocentric shuttlecock detection datasets, we introduce a dataset of 20,510 semi-automatically annotated frames captured across 11 distinct backgrounds in diverse indoor and outdoor environments, and categorize each frame into one of three difficulty levels. For labeling, we present a novel semi-automatic annotation pipeline, that enables efficient labeling from stationary camera footage. We propose a metric suited to our downstream use case and fine-tune a YOLOv8 network optimized for real-time shuttlecock detection, achieving an F1-score of 0.86 under our metric in test environments similar to training, and 0.70 in entirely unseen environments. Our analysis reveals that detection performance is critically dependent on shuttlecock size and background texture complexity. Qualitative experiments confirm their applicability to robots with moving cameras. Unlike prior work with stationary camera setups, our detector is specifically designed for the egocentric, dynamic viewpoints of mobile robots, providing a foundational building block for downstream tasks, including tracking, trajectory estimation, and system (re)-initialization.
\end{abstract}


\section{INTRODUCTION}
Robotic systems for interactive ball sports represent a compelling research frontier, requiring advances in autonomous decision-making, real-time perception, and human-robot interaction within highly dynamic environments. Badminton is particularly demanding due to its fast-paced nature, with the shuttlecock reaching extremely high velocities. Previous work by Ma et al.~\cite{Ma2025} demonstrated that accurate detection and tracking are crucial for robot performance, with limitations in these capabilities being among the factors restricting current applications to non-adversarial gameplay. 
Hence, in this work, we present a framework for automatically labeling images and demonstrate the benefits of this pipeline by fine-tuning a YOLOv8 model for one-shot shuttlecock detection.

A reliable and accurate detection framework serves as a fundamental building block for downstream applications, including tracking, trajectory estimation, zoom lens targeting, initialization, and recovery. However, existing methods assume static cameras and are evaluated in fixed viewpoints or in cropped regions~\cite{Cao2021, Lai2025}, making them unsuitable for mobile robotic platforms. The only openly available dataset~\cite{Sun2020} lacks the perspective and resolution needed for robot-mounted cameras.

\textbf{Contributions}: Existing shuttlecock detection methods are poorly suited to moving, egocentric viewpoints on mobile robots. We address this gap with a new dataset, a processing framework, and a fine-tuned detection model. Our main contributions are:
\begin{itemize}
    \item A dataset comprising 20,510 frames from badminton rallies in 11 locations,
    \item A novel annotation pipeline that achieves 85.7\% labeling accuracy through background subtraction, opponent segmentation, and temporal filtering.
    \item A fine-tuned, YOLO model for shuttlecock detection that generalizes from stationary to moving camera configurations.
\end{itemize}
All available open-source through our project website\footnote{\url{https://sites.google.com/leggedrobotics.com/shuttlecockfinder}}.
\begin{figure}[t]
    \centering
    \resizebox{\columnwidth}{!}{\import{plots/}{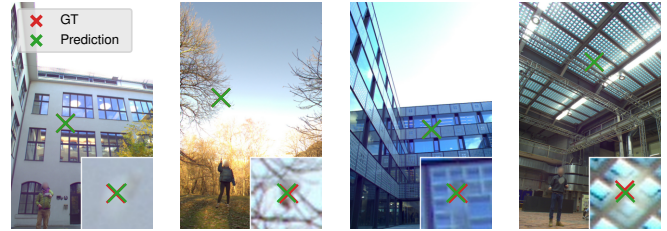}}
\caption{\textbf{Example detections of the fine-tuned model.} Our method reliably detects shuttlecocks, even under challenging conditions.}
  \label{fig:HIGHLIGHT}
\end{figure}

\section{RELATED WORK}

\subsection{Small Ball Sport Multi-Frame Trackers}
Tracking small objects, particularly balls in sports, is an active field of research. The TrackNet architecture~\cite{Huang2019,Sun2020,Chen2023,Raj2025} uses a U-Net~\cite{Ronneberger2015} to process three consecutive frames from a static camera, producing an 8-bit heatmap of ball position probabilities. Sun et al.~\cite{Sun2020} provide a dataset of 55,563 labeled frames from 18 match videos. However, all viewing angles in the dataset are from a broadcast perspective, which limits the applicability to onboard detection.
Subsequent works~\cite{Tarashima2023, Voeikov2020} extend this idea with improved heatmap representations and multi-task architectures. All these methods rely on temporal coherence between consecutive frames and are challenging to apply for fast-moving onboard cameras.

\begin{figure*}[htpb]
  \centering
  \resizebox{\textwidth}{!}{\import{plots/}{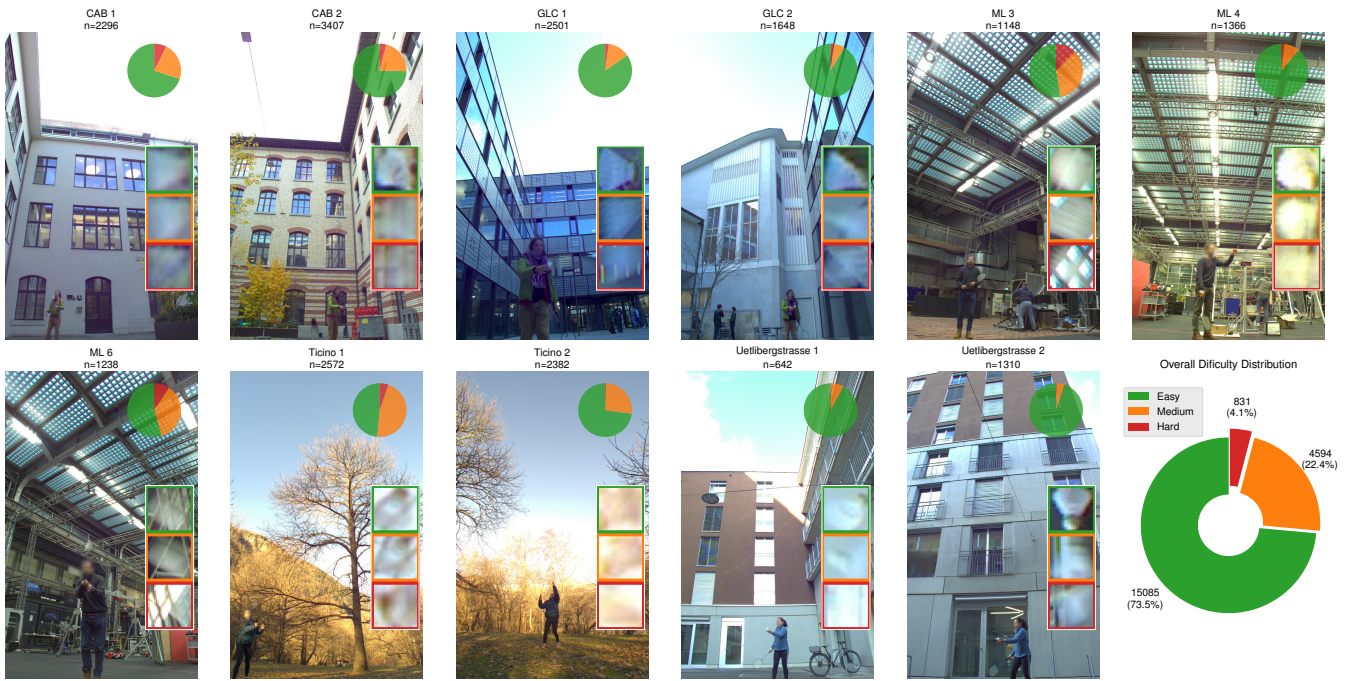}}
  \caption{\textbf{Dataset Overview and Difficulty Distribution:} Representative frames from 11 backgrounds across 5 locations. Each frame shows the per-background difficulty distribution (top-right inset) with cropped shuttlecock examples (lower-right inset), both color-coded by difficulty: green (\textit{easy}), orange (\textit{medium}), red (\textit{hard}). Sample sizes range from 642 to 3,407 frames per background. Bottom-right: Overall distribution across all 20,510 frames.}
  \label{fig:location_grid}
\end{figure*}

\subsection{Small Ball Sport Tracking-by-Detection}
Besides multi-frame trackers, several works have explored single-frame detection architectures for small object tracking in sports.
Cao et al.~\cite{Cao2021} customized the Tiny YOLOv2 architecture to detect shuttlecocks for a badminton robot application. They collected a dataset of approximately 18,000 frames, which is not publicly available.
Lai et al.~\cite{Lai2025} present YO-CSA-T, a tracking system for a robot playing badminton. They propose a static stereo camera setup and enhance YOLOv8 by adding spatial attention mechanisms, along with a trajectory estimation network to predict the future trajectory of the shuttlecock at 130 \ac{fps}. They collected a dataset with 32,539 frames, which is not publicly available.\looseness-1

While existing approaches show promising results, their datasets and models are not publicly available, and all rely on static, external camera setups. In contrast, our work provides an open dataset and model for egocentric single-frame detection, demonstrating generalization from stationary training footage to moving cameras.


\section{METHODOLOGY}

\subsection{Dataset}

We collected a dataset of 20,510 frames from 11 distinct backgrounds in various indoor, urban, and outdoor environments. \Cref{fig:location_grid} presents representative frames of each background. All images were acquired using a Basler acA1920-144uc industrial camera equipped with an \SI{8}{\milli\meter} fixed focal length lens. Rally sequences were recorded at 60 \ac{fps} with an image resolution of 1920$\times$1200 pixels. 

To characterize the challenges in shuttlecock detection, we subjectively categorized each shuttlecock instance according to three difficulty classes, with \Cref{fig:location_grid} presenting examples of these levels across all backgrounds, as well as the overall distribution and the distribution for each background. 
\begin{leftitemize}
    \item \textbf{Easy}: Shuttlecocks clearly visible and distinguishable to human annotators.
    \item \textbf{Medium}: Shuttlecocks that are barely perceptible to annotators due to significant motion blur, adverse lighting, partial occlusion, or noisy background.
    \item \textbf{Hard}: Instances where the shuttlecock is imperceptible to annotators when viewed in isolation, necessitating temporal context from adjacent frames for identification.
\end{leftitemize}

All images are annotated with a bounding box of the shuttlecock. Frames in which the shuttlecock was not visible were excluded from the dataset. For efficient labeling, we developed an automated labeling pipeline that exploits the stationary camera setup to isolate moving foreground objects from the static background, with the following steps:

\begin{leftenumerate}
    \item \textbf{Background Subtraction:} A \ac{gmm}-based algorithm~\cite{zivkovic2004} segments foreground regions from stationary camera footage. To remove noise, fill gaps, and smooth boundaries, we thereafter apply opening and closing morphological operations.

    \item \textbf{Opponent Removal:} We segment the opponent using YOLOv8-seg~\cite{Jocher2023} and remove all connected components that intersect this mask.

    \item \textbf{Pedestrian Filtering:} Detections below a predefined vertical threshold are excluded to filter pedestrians that appear too small in the image to be detected by the segmentation network.

    \item \textbf{Candidate Selection:} The remaining candidates are ranked by temporal consistency with previous detections and blob area.
\end{leftenumerate}

\begin{figure}[t]
  \centering
  \resizebox{\columnwidth}{!}{\import{plots/}{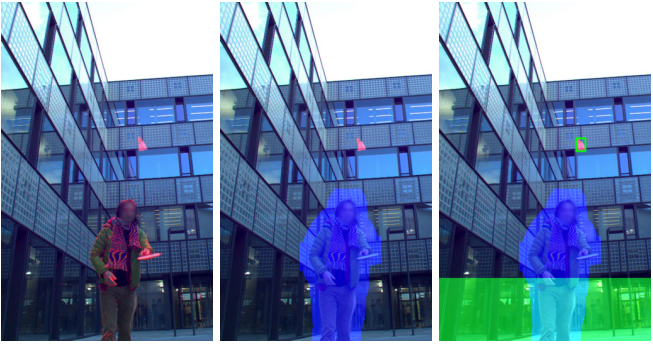}}
  \caption{\textbf{Automated Shuttlecock Labeling Pipeline:} Left: \ac{gmm}-based background segmentation identifies foreground regions (red). Center: YOLOv8-seg detects and segments the opponent player (blue), whose region is subsequently excluded from shuttlecock candidates. Right: Final detection result after applying morphological operations, person removal, and spatial constraints (lower region shown in green represents the excluded zone), with the detected shuttlecock marked by a green bounding box.}
  \label{fig:automatic_labeling}
\end{figure}

\Cref{fig:automatic_labeling} illustrates the annotation pipeline.
This pipeline correctly labeled 85.7\% of frames, with 8.3\% requiring minor bounding box adjustments and 5.9\% (primarily frames near the opponent's return stroke) requiring manual correction. All labels were reviewed to ensure quality.

\subsection{Metric}
Standard metrics based on \ac{iou} do not capture the performance characteristics most relevant to our application, where the critical requirement is the estimation of the shuttlecock's center position for downstream tasks. Therefore, we propose a distance-based evaluation scheme similar to \ac{dpr} used for tracking small objects. A detection is considered \acp{tp} if the Euclidean distance between the center of the predicted and ground truth bounding box is within $\tau = 25$ pixels. Detections are \acp{fp} if no ground truth exists or the distance exceeds $\tau$, and \acp{fn} as missed detections.
Given these definitions, we compute the precision, recall, and F1-score to evaluate our detector's performance.
We retain bounding box annotations for training, as YOLO's architecture requires them, and they enable the size-dependent analysis in Section~\ref{sec:error_analysis}.

\subsection{Training Setup}
We use a small YOLOv8 network architecture (hereafter: YOLOv8)~\cite{Jocher2023} for shuttlecock detection, where we constrain \ac{nms} to output at most one detection per frame, which is appropriate for match play and robotic demonstrations where only a single shuttlecock is used. Since all dataset frames contain a shuttlecock, 1000 background images from the COCO dataset~\cite{Lin2014} are added to the training set to expose the model to shuttlecock-free scenes and reduce \acp{fp}. To mitigate the effect of noisy labels, only \textit{easy} and \textit{medium} difficulty samples are used for training, comprising 95.9\% of the dataset. An input resolution of 1024 pixels, maintaining the aspect ratio of the original image, is chosen as a trade-off between detection accuracy and inference speed, based on a sweep from 640 to 1408 pixels. The justification for these choices is provided in the supplementary material. To further improve model generalization, we employed standard data augmentation techniques (mosaic, translation, scale, HSV, mixup, and flipping).  Among these, mixup augmentation~\cite{zhang2017} improved performance most significantly, increasing the overall recall from 0.68 to 0.78 while maintaining the same precision.

\section{EVALUATION}

We evaluate the trained model in three ways. First, we quantify performance on stationary camera footage using location-based and background-based cross-validation to assess generalization to unseen environments. Second, we analyze the primary sources of detection error. Third, we present qualitative results with a moving camera to validate the applicability to mobile robotic platforms.

\subsection{Quantitative Results with Stationary Camera}
\label{sec:quantitative_results}

During the fine-tuning phase, it became evident that model performance depended on the background and location characteristics, which manifested as high variance across individual subsets in the validation metrics. To systematically assess the model's generalization capabilities and mitigate overfitting to specific environmental conditions, we conducted the following two experiments with fixed hyper parameters:
\begin{itemize}
   \item \textbf{Background-based cross-validation}: Eleven runs, each withholding a single background while training on all others (including same-location backgrounds) to evaluate generalization to similar but distinct environments.
    \item \textbf{Location-based cross-validation}: Five runs, each withholding all backgrounds from one location for testing to evaluate generalization to entirely unseen environments.
\end{itemize}

\begin{figure}[t]
  \centering
  \resizebox{\columnwidth}{!}{\input{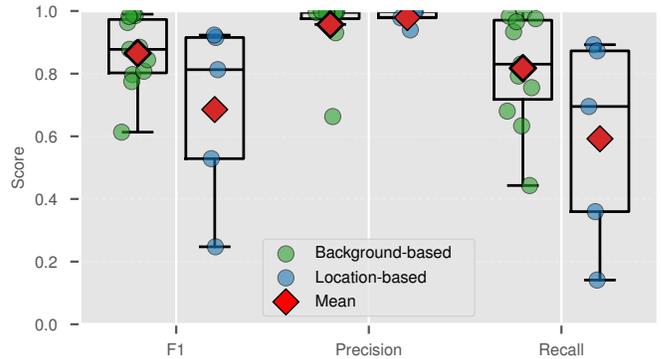}}
  \caption{
      \textbf{Cross-Validation Results Distribution:} Performance metrics across individual backgrounds (left/green) and locations (right/blue). Each circle represents model performance when trained on all subsets except one and evaluated on the held-out subset. Background-based validation indicates how well the model generalizes to environments similar to those in the training set, while location-based validation indicates generalization to previously unseen environments. Diamonds indicate mean performance across all folds.
  }
  \label{fig:crossval_combined}
\end{figure}

\Cref{fig:crossval_combined} presents the results of both cross-validation strategies. In background-based validation, the model achieves a mean precision of 0.957 with a recall ranging from 0.443 to 0.999, demonstrating a strong generalization to environments similar to those in training. A single precision outlier corresponds to the \texttt{GLC\_2} subset, where a static background feature is consistently misclassified as a shuttlecock, particularly in difficult samples where the model assigns low confidence to the actual shuttlecock. A downstream trajectory estimation module must account for this failure mode and filter it out. Location-based validation reveals higher variance, with recall ranging from 0.141 to 0.892: the model generalizes effectively across urban locations (\texttt{GLC}, \texttt{CAB}, \texttt{Uetlibergstrasse}) but poorly to unseen environments such as \texttt{ML} and \texttt{Ticino}, indicating that deployment in non-urban settings requires additional data collection. The precision remains consistently high at all locations. Detailed numerical results for both cross-validation experiments are provided in the supplementary material.

\Cref{tab:cv_results} presents the general results of both evaluation strategies and the results for each difficulty level. The general results differ slightly from the mean values shown in \Cref{fig:crossval_combined}, since the results here represent subset-size-weighted accumulations rather than unweighted means. As expected, performance decreases with increasing difficulty level. Notably, recall drops substantially at harder difficulty levels while precision remains high.

\begin{table}
\centering
\caption{Location-wise vs.\ Background-wise cross-validation by difficulty. Values are \textit{Loc / Bg}.}
\label{tab:cv_results}
\resizebox{1\columnwidth}{!}{
\setlength{\tabcolsep}{6pt}
\begin{tabular}{lcccc}
\toprule
\textbf{Metric} & \textbf{Overall} & \textbf{Easy} & \textbf{Medium} & \textbf{Hard} \\
\midrule
F1        & 0.703 / 0.864 & 0.788 / 0.920 & 0.414 / 0.688 & 0.238 / 0.576 \\
Precision & 0.970 / 0.954 & 0.977 / 0.961 & 0.934 / 0.943 & 0.772 / 0.810 \\
Recall    & 0.552 / 0.789 & 0.661 / 0.882 & 0.266 / 0.542 & 0.140 / 0.447 \\
\bottomrule
\end{tabular}
}
\end{table}

Finally, we analyze the effect of the distance threshold $\tau$ on detection performance. For true positives, the mean Euclidean distance between the predicted bounding box center and the ground truth center was less than 1~pixel in both test cases ($0.89$~pixels for background-based and $0.97$~pixels for location-based), indicating that correct detections exhibit minimal offset and there is negligible dependence on $\tau$.

\subsection{Error Analysis}
\label{sec:error_analysis}

\begin{figure}[h]
  \centering
  \resizebox{\columnwidth}{!}{\input{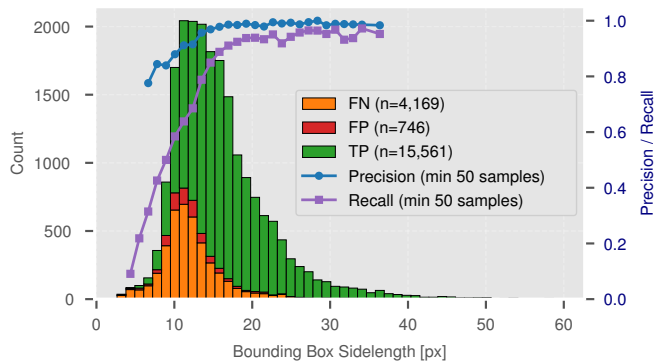}}
  \caption{
\textbf{Distribution of Bounding Box Sizes and Corresponding Model Performance:} The histogram shows the distribution of correct (green, n=15,561) and incorrect (red, n=4,915) predictions across different bounding box side lengths measured in input pixels and defined as the geometric mean $\sqrt{w \cdot h}$. The blue line represents model precision, the violet line represents recall (both calculated for bins with at least 50 samples), plotted on the secondary y-axis on the right.}

  \label{fig:size_acc_distribution}
\end{figure}

A critical factor affecting detection performance is the size of the shuttlecock in the input image. \Cref{fig:size_acc_distribution} presents the shuttlecock size distribution in our dataset along with the corresponding detection performance. The results demonstrate a strong size-dependent performance characteristic: below a bounding box side length of approximately 20 pixels, recall begins to drop, and below 15 pixels, precision also degrades. Above the 20-pixel threshold, recall plateaus above 90\% and precision reaches nearly 100\%, indicating that further increases in shuttlecock size provide marginal performance gains. The majority of samples from our dataset are concentrated in the 10-20 pixel range, where performance transitions from poor to excellent.

\subsection{Qualitative Results with Moving Camera}

\begin{figure}[t]
    \centering
    \resizebox{\columnwidth}{!}{\import{plots/}{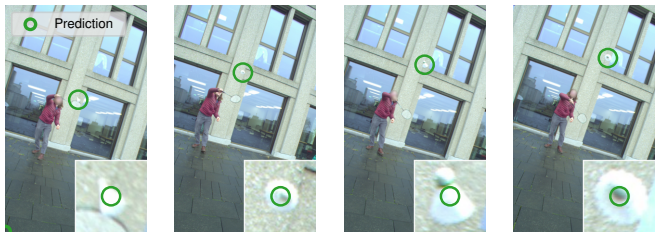}}
\caption{\textbf{Predictions with moving camera:} Example sequence from \texttt{LEE\_moving\_1} showing correct shuttlecock detection across evenly sampled frames from a sequence of 60 frames.}
  \label{fig:moving_pred}
\end{figure}

We conducted a qualitative analysis using a moving camera across three different scenes (video sequences in the supplementary material). In rallies \texttt{LEE\_moving\_1} and \texttt{LEE\_moving\_2}, where the opponent was close and the background was uniform, detection accuracy remained high throughout. Samples of the predictions are shown in \Cref{fig:moving_pred}. Conversely, the \texttt{Ticino\_moving\_1} sequence, characterized by increased background clutter and the opponent being farther from the camera, demonstrated reduced detection reliability, with consistent detection achieved primarily when the shuttlecock was silhouetted against the sky.

Despite these environmental dependencies, the results suggest that the proposed detection framework generalizes to dynamic camera configurations, supporting its applicability to robotic badminton systems with mobile vision platforms.

\section{CONCLUSIONS}

We presented a one-shot shuttlecock detection system comprising a multi-environment dataset, an automated labeling pipeline, and a fine-tuned YOLOv8 detector. The detector generalizes well to environments similar to training, but exhibits reduced performance in entirely unseen locations with complex backgrounds; qualitative experiments confirm applicability to moving cameras on legged robots. Future work should focus on two key directions to enhance the robustness of the system. Expanding the dataset to include more diverse environments would improve generalization, and our automated labeling approach would enable efficient, scalable data collection. Exploring architectural modifications inspired by related work, such as multi-frame inputs or attention mechanisms, may further improve detection accuracy on small and distant shuttlecocks.

\addtolength{\textheight}{-4.3cm}   


\begin{table*}[h]
\centering
\caption{Supplementary Material}
\label{tab:supplementary_material}
\begin{tabularx}{\textwidth}{l X p{4.5cm}}
    \toprule
    \textbf{Material} & \textbf{Description} & \textbf{Link} \\
    \midrule
    Dataset 
        & Dataset in Ultralytics YOLO format, with separate subsets for each recording location and difficulty level, provided over Google Suite. 
        & \url{https://drive.google.com/drive/folders/1neBfC7-Fp-l53muxZiIbIET6lYH5xv9s?usp=sharing}\\
    \midrule
    Code 
        & Repository containing the best model and code described in this paper 
        & \url{https://github.com/leggedrobotics/shuttle_detection} \\
    \midrule
    \multirow[t]{6}{*}{Additional Results} 
        & \textbf{Table I:} Adding COCO backgrounds to both training and validation. 
        & \multirow[t]{6}{4.5cm}{\url{https://drive.google.com/file/d/1vf-zJh2ijhvdCb6varUMPmfYpV3pkxiV/view?usp=drive_link}} \\
    \cmidrule{2-2}
        & \textbf{Table II:} Validation metrics by difficulty level. 
        & \\
    \cmidrule{2-2}
        & \textbf{Table III:} Location-wise cross-validation results. 
        & \\
    \cmidrule{2-2}
        & \textbf{Table IV:} Background-wise cross-validation results. 
        & \\
    \cmidrule{2-2}
        & \textbf{Figure 1:} Model performance and inference time vs. input image size. 
        & \\
    \cmidrule{2-2}
        & \textbf{Figure 2:} Precision-recall curves. 
        & \\
    \midrule
    \multirow[t]{3}{*}{Videos} 
        & \texttt{LEE\_moving\_1} (00:23): Shuttlecock detection with a moving camera in an urban setting with a heterogeneous background. 
        & \url{https://youtube.com/shorts/JM6ZTMC9Z9U?feature=share} \\
    \cmidrule{2-3}
        & \texttt{LEE\_moving\_2} (00:47): Shuttlecock detection with a moving camera in an urban setting with a homogeneous background. 
        & \url{https://youtube.com/shorts/nzEzU3-Lv1s?feature=share} \\
    \cmidrule{2-3}
        & \texttt{Ticino\_moving\_1} (00:28): Shuttlecock detection with a moving camera in an outdoor setting. 
        & \url{https://youtube.com/shorts/5UPuVgabico?feature=share} \\
    \bottomrule
\end{tabularx}
\end{table*}

{
\linespread{0.95}
\bibliographystyle{IEEEtranBST/IEEEtran}
\bibliography{IEEEtranBST/IEEEabrv,IEEEtranBST/references}

\begin{thebibliography}{10}
\providecommand{\url}[1]{#1}
\csname url@rmstyle\endcsname
\providecommand{\newblock}{\relax}
\providecommand{\bibinfo}[2]{#2}
\providecommand\BIBentrySTDinterwordspacing{\spaceskip=0pt\relax}
\providecommand\BIBentryALTinterwordstretchfactor{4}
\providecommand\BIBentryALTinterwordspacing{\spaceskip=\fontdimen2\font plus
\BIBentryALTinterwordstretchfactor\fontdimen3\font minus \fontdimen4\font\relax}
\providecommand\BIBforeignlanguage[2]{{%
\expandafter\ifx\csname l@#1\endcsname\relax
\typeout{** WARNING: IEEEtran.bst: No hyphenation pattern has been}%
\typeout{** loaded for the language `#1'. Using the pattern for}%
\typeout{** the default language instead.}%
\else
\language=\csname l@#1\endcsname
\fi
#2}}

\bibitem{Ma2025}
Y.~Ma, A.~Cramariuc, F.~Farshidian, and M.~Hutter, ``Learning coordinated badminton skills for legged manipulators,'' \emph{Science robotics}, vol.~10, no. 102, p. eadu3922, 2025.

\bibitem{Cao2021}
\BIBentryALTinterwordspacing
Z.~Cao, T.~Liao, W.~Song, Z.~Chen, and C.~Li, ``Detecting the shuttlecock for a badminton robot: A yolo based approach,'' \emph{Expert Systems with Applications}, vol. 164, p. 113833, 2 2021. [Online]. Available: \url{https://www.sciencedirect.com/science/article/pii/S0957417420306436}
\BIBentrySTDinterwordspacing

\bibitem{Lai2025}
Y.~Lai, Z.~Shi, and C.~Zhu, ``Yo-csa-t: A real-time badminton tracking system utilizing yolo based on contextual and spatial attention,'' in \emph{2025 IEEE/RSJ International Conference on Intelligent Robots and Systems (IROS)}, 2025, pp. 732--739.

\bibitem{Sun2020}
N.~E. Sun, Y.~C. Lin, S.~P. Chuang, T.~H. Hsu, D.~R. Yu, H.~Y. Chung, and T.~U. Ik, ``Tracknetv2: Efficient shuttlecock tracking network,'' in \emph{Proceedings - 2020 International Conference on Pervasive Artificial Intelligence, ICPAI 2020}.\hskip 1em plus 0.5em minus 0.4em\relax Institute of Electrical and Electronics Engineers Inc., 12 2020, pp. 86--91.

\bibitem{Huang2019}
Y.-C. Huang, I.-N. Liao, C.-H. Chen, T.-U. İk, and W.-C. Peng, ``Tracknet: A deep learning network for tracking high-speed and tiny objects in sports applications,'' in \emph{2019 16th IEEE International Conference on Advanced Video and Signal Based Surveillance (AVSS)}, 2019, pp. 1--8.

\bibitem{Chen2023}
Y.~J. Chen and Y.~S. Wang, ``Tracknetv3: Enhancing shuttlecock tracking with augmentations and trajectory rectification,'' in \emph{Proceedings of the 5th ACM International Conference on Multimedia in Asia, MMAsia 2023}.\hskip 1em plus 0.5em minus 0.4em\relax Association for Computing Machinery, Inc, 12 2023.

\bibitem{Raj2025}
A.~Raj, L.~Wang, and T.~Gedeon, ``Tracknetv4: Enhancing fast sports object tracking with motion attention maps,'' in \emph{ICASSP, IEEE International Conference on Acoustics, Speech and Signal Processing - Proceedings}.\hskip 1em plus 0.5em minus 0.4em\relax Institute of Electrical and Electronics Engineers Inc., 2025.

\bibitem{Ronneberger2015}
\BIBentryALTinterwordspacing
O.~Ronneberger, P.~Fischer, and T.~Brox, ``U-net: Convolutional networks for biomedical image segmentation,'' \emph{Lecture Notes in Computer Science (including subseries Lecture Notes in Artificial Intelligence and Lecture Notes in Bioinformatics)}, vol. 9351, pp. 234--241, 2015. [Online]. Available: \url{https://link.springer.com/chapter/10.1007/978-3-319-24574-4_28}
\BIBentrySTDinterwordspacing

\bibitem{Tarashima2023}
\BIBentryALTinterwordspacing
S.~Tarashima, M.~A. Haq, Y.~Wang, and N.~Tagawa, ``Widely applicable strong baseline for sports ball detection and tracking,'' 2023. [Online]. Available: \url{https://arxiv.org/abs/2311.05237}
\BIBentrySTDinterwordspacing

\bibitem{Voeikov2020}
\BIBentryALTinterwordspacing
R.~Voeikov, N.~Falaleev, and R.~Baikulov, ``Ttnet: Real-time temporal and spatial video analysis of table tennis,'' \emph{IEEE Computer Society Conference on Computer Vision and Pattern Recognition Workshops}, vol. 2020-June, pp. 3866--3874, 4 2020. [Online]. Available: \url{https://arxiv.org/pdf/2004.09927}
\BIBentrySTDinterwordspacing

\bibitem{zivkovic2004}
\BIBentryALTinterwordspacing
Z.~Zivkovic, ``Improved adaptive gaussian mixture model for background subtraction,'' \emph{Proceedings - International Conference on Pattern Recognition}, vol.~2, pp. 28--31, 2004. [Online]. Available: \url{https://ieeexplore.ieee.org/document/1333992}
\BIBentrySTDinterwordspacing

\bibitem{Jocher2023}
\BIBentryALTinterwordspacing
G.~Jocher, J.~Qiu, and A.~Chaurasia, ``Ultralytics yolo,'' 1 2023. [Online]. Available: \url{https://github.com/ultralytics/ultralytics}
\BIBentrySTDinterwordspacing

\bibitem{Lin2014}
T.-Y. Lin, M.~Maire, S.~Belongie, J.~Hays, P.~Perona, D.~Ramanan, P.~Doll{\'a}r, and C.~L. Zitnick, ``Microsoft coco: Common objects in context,'' in \emph{European conference on computer vision}.\hskip 1em plus 0.5em minus 0.4em\relax Springer, 2014, pp. 740--755.

\bibitem{zhang2017}
\BIBentryALTinterwordspacing
H.~Zhang, M.~Cisse, Y.~N. Dauphin, and D.~Lopez-Paz, ``mixup: Beyond empirical risk minimization,'' 2018. [Online]. Available: \url{https://arxiv.org/abs/1710.09412}
\BIBentrySTDinterwordspacing

\end{thebibliography}
}
\end{document}